\documentclass[conference]{IEEEtran}

\usepackage{graphicx}
\usepackage{amsmath,amssymb}
\usepackage{booktabs}
\usepackage{multirow}
\usepackage{cite}
\usepackage{url}
\usepackage{xcolor}
\usepackage[hidelinks]{hyperref}

\usepackage[caption=false,font=footnotesize]{subfig}

\usepackage{xurl}

\newcommand{\method}{LiteKD-Net}

\newcommand{\etal}{et al.\ }

\title{LiteKD-Net: Lightweight Knowledge-Distilled Network for Mobile Image Denoising}

\author{
\IEEEauthorblockN{Zhiyi Zhou}
\IEEEauthorblockA{
Global College\\
Shanghai Jiaotong University\\
Shanghai, China\\
\{georgeyy\}@sjtu.edu.cn
}
}

\begin{document}

\maketitle


\begin{abstract}
Mobile image denoising requires both good restoration quality and
low computational cost. In addition, it's annoying to collect large-scale LQ-GT clean pairs.
As a result, we propose \method, a lightweight
knowledge-distilled network for mobile image denoising. First, a
physics-guided noise simulation pipeline generates paired training data by adding pixel crosstalk compared with pipelines applied to cameras. Next, we adapt the Real-ESRGAN to identity-resolution
denoising and construct a
lightweight Student using Lite-RRDB blocks based on depthwise separable
convolutions. Third, feature-level knowledge
distillation is applied to transfer the Teacher's restoration
capability to the Student without introducing additional inference
cost. Experiments on real-world datasets show that our model reaches great reduction in runtime and increase in the inference rate with good restoration quality. Our model also reaches the best in all metrics compared with SwinIR. These results indicate that LiteKD-Net provides a great trade-off between restoration quality and computational efficiency.
\end{abstract}

\begin{IEEEkeywords}
computer vision, deep learning, image restoration, image denoising, lightweight model
\end{IEEEkeywords}

\vspace{-8pt}
\section{Introduction}
\vspace{-4pt}
\label{sec:introduction}

Mobile phones have become the most widely used imaging devices. Compared with professional
digital single-lens reflex cameras, mobile devices employ substantially smaller
image sensors and smaller individual pixels, which capture fewer photons
per exposure and therefore produce intrinsically lower
signal-to-noise ratios~\cite{abdelhamed2018sidd}. Increasing sensor
gain can amplify weak captured signal but simultaneously increase  noise, banding, and
other sensor-specific degradations~\cite{wei2020physics}. Effective
denoising is therefore critical for mobile image restoration.

Deep convolutional neural networks have substantially advanced image
denoising performance. However, many high-capacity restoration networks contain tens of millions of parameters and require billions of floating-point
operations. When deployed on mobile
hardware, these models can suffer from high latency, excessive energy
consumption, and unacceptable memory footprints. Cloud-based online processing reduces local computation but introduces communication latency, bandwidth costs, and user privacy concerns. Mobile image denoising therefore demands a model that optimizes restoration accuracy and on-device deployment efficiency.

A second practical challenge is the limited availability of strictly
aligned image pairs. Since most denoising works focus on the professional cameras, simulation pipeline for mobile imaging is not widely researched. Capturing the
same scene with identical content but different noise characteristics
is difficult because subject motion, illumination changes, and
in-camera processing pipelines can introduce spatial misalignment.
Purely synthetic additive Gaussian noise is convenient but does not
adequately represent the signal: dependent and structured noise produced by real CMOS
sensors~\cite{wei2020physics,abdelhamed2018sidd}.

In this work, we propose a unified framework for mobile image denoising
that integrates three components. First, a
\textbf{Noise Simulation Pipeline (NSP)} constructs aligned
GT and LQ training data by adding pixel crosstalk based on the professional DSLR cameras' simulation pipeline~\cite{wei2020physics}. Second, a
\textbf{Lite-RRDB} architecture replaces standard RRDB blocks with
lightweight residual dense blocks built from depthwise separable
convolutions~\cite{howard2017mobilenets}, reducing parameter count
and computational cost greatly based on Real-ESRGAN~\cite{wang2021realesrgan}. Third,
\textbf{Noise-KD} transfers restoration capability from a frozen high-capacity Teacher~\cite{wang2021realesrgan} to the lightweight Student through
feature-level knowledge distillation~\cite{hinton2015distilling},
recovering restoration quality without increasing inference cost.

Our main contributions are summarized as follows:
\begin{itemize}
    \item We design a \textbf{Noise Simulation Pipeline (NSP)} that
    generates perfectly aligned GT and LQ data.

    \item We construct a \textbf{Lite-RRDB} lightweight module in which original RRDB blocks are replaced by
    depthwise separable residual dense blocks.

    \item We introduce \textbf{Noise-KD}, a feature-level knowledge
    distillation strategy that transfers restoration
    capability from a frozen Heavy Teacher to the Student.

    \item We evaluate the model on real-world datasets. Ablation study and comparison experiment all show that our model has great denoising performance with smaller model size and faster inference speed.
\end{itemize}

\begin{figure}[!t]
    \centering
    \includegraphics[width=\columnwidth]
    {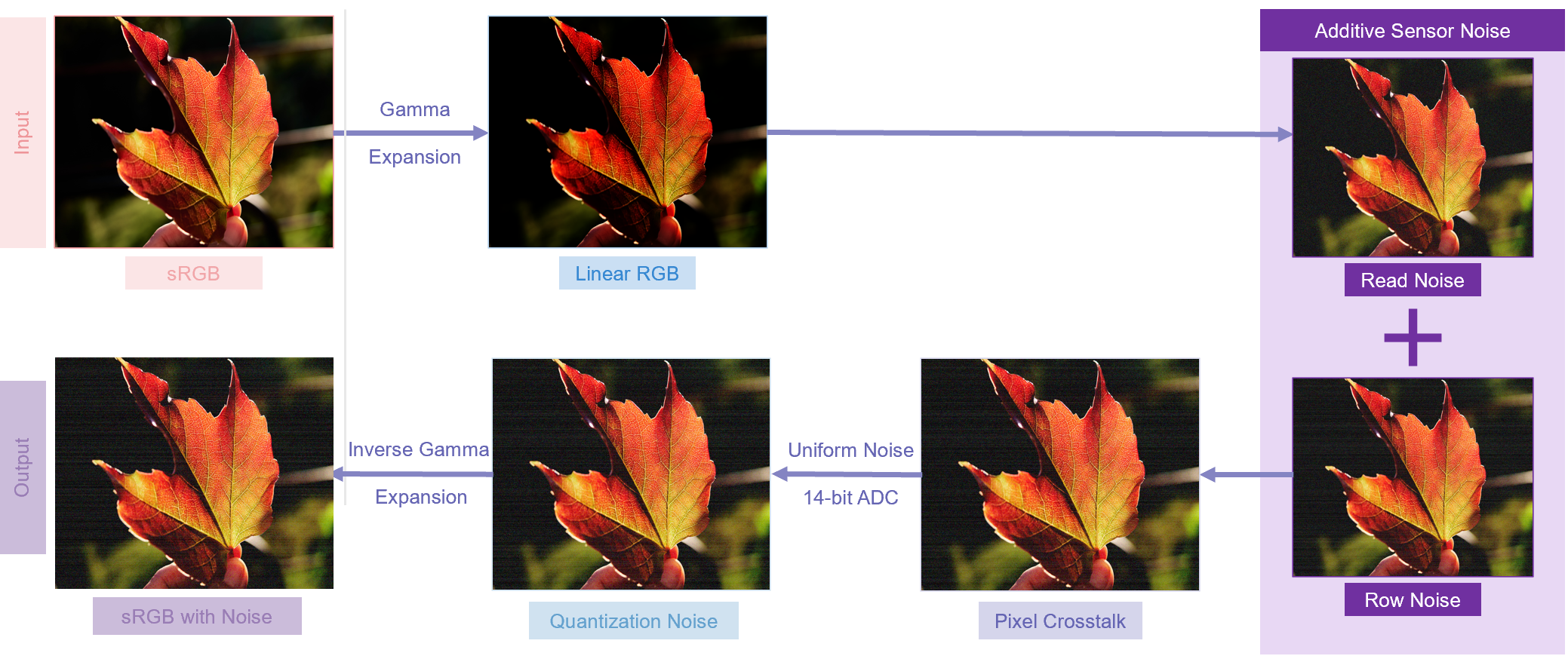}
    \vspace{-16pt}
    \caption{
    Overview of NSP.
    A clean sRGB image is first approximately linearized. Photon shot
    noise, Tukey--Lambda read noise, and row noise are then simulated
    following Wei \etal~\cite{wei2020physics}. Our proposed pixel
    crosstalk introduces local spatial correlation between neighboring
    pixels, after which quantization noise is added. The result is
    clipped and converted back to sRGB.
    }
    \label{fig:nsp_pipeline}
    \vspace{-16pt}
\end{figure}
\vspace{-6pt}
\section{Related Work}
\label{sec:related_work}
\vspace{-2pt}

\subsection{Physics-Based Noise Modeling.}
Real CMOS noise is more complex than additive white Gaussian noise.
Foi \etal~\cite{foi2008practical} established the Poisson--Gaussian
model. Wei \etal~\cite{wei2020physics}
proposed a physics-based formation model decomposing RAW noise into
photon shot noise, Tukey--Lambda read noise, row noise, and quantization
noise. CBDNet~\cite{guo2019cbdnet} combined realistic noise synthesis
with real photographs for blind denoising, while SIDD~\cite{abdelhamed2018sidd}
and MIDD~\cite{flepp2024midd} provide large-scale smartphone
denoising dataset.

\subsection{Image Denoising.}
DnCNN~\cite{zhang2017beyond} pioneered residual learning for blind
denoising; RIDNet~\cite{anwar2019ridnet} advanced real-image denoising
with feature attention. RRDB-based generators~\cite{wang2021realesrgan} provide strong restoration backbones. For efficiency, depthwise separable
convolution~\cite{howard2017mobilenets} reduces computation;
LPIENet~\cite{conde2023lpienet} and MOFA~\cite{chen2023mofa} target
on-device restoration.

\section{Method}
\label{sec:method}

\subsection{Problem Formulation}

Let
\vspace{-2pt}
\begin{equation}
\mathcal{D}=\{(\mathbf{x}_i,\mathbf{y}_i)\}_{i=1}^{N}
\end{equation}
\vspace{-2pt}
denote a paired denoising dataset, where
$\mathbf{x}_i\in[0,1]^{H\times W\times3}$ is a noisy RGB image and
$\mathbf{y}_i$ is the corresponding clean image. The lightweight
Student network $f_S(\cdot;\theta_S)$ predicts
\vspace{-2pt}
\begin{equation}
\hat{\mathbf{y}}_i
=
f_S(\mathbf{x}_i;\theta_S).
\label{eq:student_output}
\end{equation}
\vspace{-2pt}
The objective is to restore a visually clean image while preserving
details. At the same time, the Student
must reduce parameter count, computational complexity, memory
consumption, and inference latency for mobile deployment.

\subsection{Proposed Approach}
\begin{figure*}[!t]
    \centering
    \includegraphics[width=\textwidth]
    {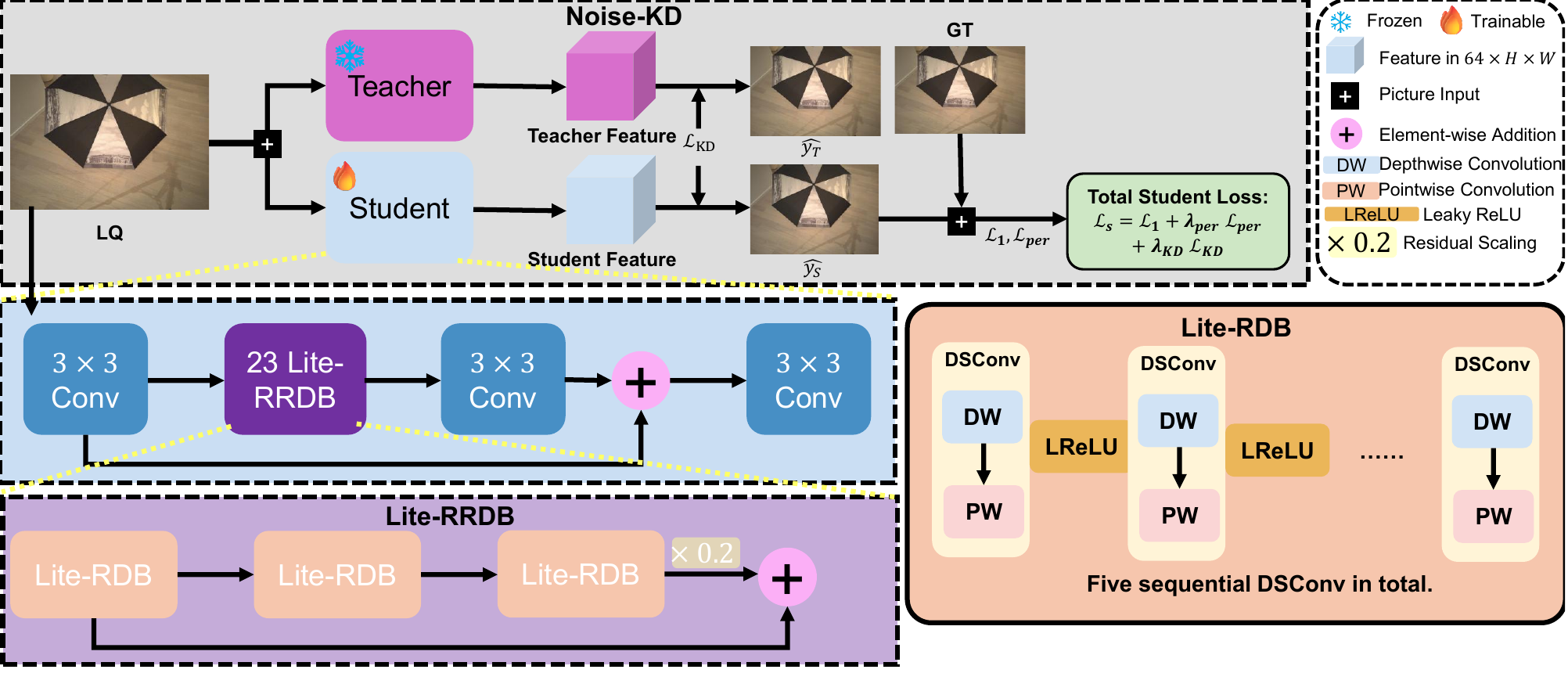}
    \vspace{-12pt}
    \caption{
    Overall process of training process.
    }
    \label{fig:overall}
    \vspace{-18pt}
\end{figure*}

The complete procedure of our simulation pipeline is shown in Fig.~\ref{fig:nsp_pipeline} and the overall training process is shown in Fig.~\ref{fig:overall}.

\subsubsection{\textbf{Noise Simulation Pipeline(NSP)}}

Collecting large-scale dataset is difficult because camera motion, scene motion, exposure variation, and
in-camera processing can introduce spatial or photometric misalignment.
We therefore simulate synthetic training pairs from clean sRGB images.

Our NSP is adapted from the model of
Wei et al.~\cite{wei2020physics}, which considers photon shot noise,
heavy-tailed read noise, row noise, and quantization noise. Since our
training images are processed sRGB images rather than RAW measurements,
we first approximately linearize each clean image using gamma expansion,
apply the sensor-noise simulation in the linear RGB domain, and finally
convert the generated image back to sRGB.

\begin{figure}[!t]
    \centering
    \includegraphics[width=\columnwidth]
    {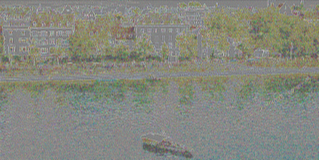}
    \vspace{-12pt}
    \caption{
    Isolated effect of pixel crosstalk. The two images use identical
samples for all other noise; only the crosstalk
kernel differs. The amplified signed difference 
reveals edge alignment and smooth residual patterns, 
confirming that the kernel introduces spatial correlation 
rather than independent noise.
    }
    \label{fig:crosstalk_difference}
    \vspace{-18pt}
\end{figure}

The main contribution of our NSP is the explicit modeling of
\textbf{pixel crosstalk}. The original formulation assumes that the
major sensor-noise components are generated without explicitly modeling
local interactions between neighboring pixels. However, in compact
mobile sensors, optical or electrical leakage may cause the measurement
of one pixel to influence its nearby pixels, producing spatially
correlated noise. 
Let
$\mathbf{s}_{\mathrm{pre}} = K\operatorname{Poisson}
\big( \frac{\Gamma(\mathbf{y}_{s})}{K} \big)
+ \mathbf{n}_{\mathrm{read}} + \mathbf{n}_{\mathrm{row}}$
denote the signal after shot, read, and row noise.
The complete NSP is:

\vspace{-2pt}
\begin{equation}
\label{eq:nsp_complete}
\hat{\mathbf{x}}_{s}
=
\Gamma^{-1}
\Big(
\operatorname{clip}
\big(
\mathbf{s}_{\mathrm{pre}} * \boldsymbol{\kappa}_{c}
+ \mathbf{n}_{q},
\; 0, \; 1
\big)
\Big).
\end{equation}
\vspace{-2pt}

where $\mathbf{y}_{s}$ and $\hat{\mathbf{x}}_{s}$ denote the clean and
simulated noisy sRGB images, respectively, and $*$ denotes channel-wise
spatial convolution. The proposed crosstalk kernel is
\vspace{-2pt}
\begin{equation}
\label{eq:crosstalk}
\boldsymbol{\kappa}_{c}
=
\begin{bmatrix}
c/4 & c & c/4 \\
c & 1-5c & c \\
c/4 & c & c/4
\end{bmatrix}.
\end{equation}
\vspace{-2pt}
Because the kernel weights sum to one, it approximately preserves local
brightness while diffusing a small fraction of each pixel signal to its
neighbors. This changes independent pixel noise into locally correlated
noise and introduces only mild spatial spreading. As shown in
Fig.~\ref{fig:crosstalk_difference}, the signed difference is mainly
concentrated around noise impulses, edges, and fine textures, directly
visualizing the effect introduced by pixel crosstalk. Detailed
definitions of the individual noise distributions and their parameters
are provided in Appendix~\ref{app:nsp_details}.

\subsubsection{\textbf{Lite-RRDB}}

We first adapt the Real-ESRGAN~\cite{wang2021realesrgan} from super-resolution to
identity-resolution denoising. Since denoising preserves the spatial
resolution, all terminal upsampling modules are removed, producing an
$H\times W \rightarrow H\times W$ Heavy Teacher. Its backbone contains
23 RRDBs, where each RRDB consists of three Residual Dense Blocks
(RDBs).

Although the Heavy Teacher provides strong restoration capacity, its
standard convolutions and dense feature concatenations result in a high
parameter count and computational cost. We therefore replace every
RRDB with a Lite-RRDB.

In an original RDB, five standard $3\times3$ convolutions are connected
through dense concatenation, such that each layer receives the outputs
of all preceding layers. In our Lite-RDB, these concatenation paths are
removed. The five layers are executed sequentially, and the feature
width remains fixed at 64 channels throughout the block.

Each standard convolution is further replaced by a depthwise separable
convolution:
\vspace{-2pt}
\begin{equation}
\label{eq:dsconv}
\operatorname{DSConv}(\mathbf{z})
=
\operatorname{PWConv}_{1\times1}
\left(
\operatorname{DWConv}_{3\times3}(\mathbf{z})
\right).
\end{equation}
\vspace{-2pt}
The depthwise convolution extracts spatial information independently
within each channel, while the pointwise convolution performs
cross-channel feature fusion at each spatial location. For a kernel
size $k$, the parameter counts of a standard convolution and a
depthwise separable convolution are
\vspace{-2pt}
\begin{equation}
\label{eq:dsconv_params}
P_{\mathrm{std}}
=
k^{2}C_{\mathrm{in}}C_{\mathrm{out}},
\qquad
P_{\mathrm{DS}}
=
k^{2}C_{\mathrm{in}}
+
C_{\mathrm{in}}C_{\mathrm{out}}.
\end{equation}
\vspace{-2pt}
Importantly, Lite-RRDB removes dense concatenation but does not remove
residual learning. Each Lite-RDB retains its local residual connection
with a residual scaling factor of $0.2$, and each Lite-RRDB block also
retains the outer residual connection across its three internal
Lite-RDBs. Therefore, Lite-RRDB preserves the residual-in-residual
training structure while substantially reducing computational costs.

\subsubsection{\textbf{Noise-KD}}
Architectural compression improves efficiency but also reduces the
representation capacity of the Student. To recover part of this lost
restoration ability, we introduce Noise-KD, feature-level knowledge
distillation.

The identity-resolution RRDB network is first trained as the Heavy
Teacher using paired noisy-clean images. After convergence, all Teacher
parameters are frozen. During Student training, the same noisy image
$\mathbf{x}$ is passed through both networks. Let
$\mathbf{F}_{T}(\mathbf{x})$ and $\mathbf{F}_{S}(\mathbf{x})$ denote the
Teacher and Student body features obtained after the corresponding body
convolution. Noise-KD minimizes their mean-squared difference:
\vspace{-2pt}
\begin{equation}
\label{eq:noise_kd}
\mathcal{L}_{\mathrm{KD}}
=
\frac{1}{CHW}
\left\|
\mathbf{F}_{S}(\mathbf{x})
-
\mathbf{F}_{T}(\mathbf{x})
\right\|_{2}^{2}.
\end{equation}
\vspace{-2pt}
The Student is jointly optimized using pixel reconstruction,
perceptual supervision, and feature distillation:
\vspace{-2pt}
\begin{equation}
\label{eq:total_loss}
\mathcal{L}_{\mathrm{total}}
=
\mathcal{L}_{1}
+
\lambda_{\mathrm{per}}
\mathcal{L}_{\mathrm{per}}
+
\lambda_{\mathrm{KD}}
\mathcal{L}_{\mathrm{KD}},
\end{equation}
\vspace{-2pt}
where $\mathcal{L}_{1}$ encourages pixel-level fidelity,
$\mathcal{L}_{\mathrm{per}}$ preserves perceptually important
structures, and $\mathcal{L}_{\mathrm{KD}}$ transfers intermediate
restoration knowledge from the Heavy Teacher.

The Teacher is required only during training. At inference time, it is
discarded, and only the Lite-RRDB Student is used. Noise-KD therefore
improves the representation of the lightweight model without adding any
inference-time parameters or computation.
\vspace{-8pt}
\subsection{Training Objective}

The Teacher is trained using a reconstruction objective and a perceptual
objective:
\vspace{-2pt}
\begin{equation}
\mathcal{L}_{T}
=
\mathcal{L}_{1}
+
\lambda_{\mathrm{per}}
\mathcal{L}_{\mathrm{LPIPS}}.
\end{equation}
\vspace{-2pt}
The Student reconstruction loss is
\vspace{-2pt}
\begin{equation}
\mathcal{L}_{\mathrm{rec}}
=
\left\|
f_S(\mathbf{x})-\mathbf{y}
\right\|_1,
\end{equation}
\vspace{-2pt}
and the perceptual loss is
\vspace{-2pt}
\begin{equation}
\mathcal{L}_{\mathrm{per}}
=
\operatorname{LPIPS}
\left(
f_S(\mathbf{x}),\mathbf{y}
\right).
\end{equation}
\vspace{-2pt}
The complete Student objective is
\vspace{-2pt}
\begin{equation}
\mathcal{L}_{\mathrm{total}}
=
\mathcal{L}_{\mathrm{rec}}
+
\lambda_{\mathrm{per}}\mathcal{L}_{\mathrm{per}}
+
\lambda_{\mathrm{KD}}\mathcal{L}_{\mathrm{KD}},
\label{eq:total_loss}
\end{equation}
\vspace{-2pt}
where $\lambda_{\mathrm{per}}$ and $\lambda_{\mathrm{KD}}$
control the contributions of perceptual supervision and feature
distillation. The Teacher remains frozen, and only the Student
parameters are updated.

\section{Experiments}
\label{sec:experiments}
\begin{table}[t]
\centering
\caption{Comparison with Real-ESRGAN~\cite{wang2021realesrgan} and
SwinIR~\cite{liang2021swinir}.
Best results are in \textcolor{red}{red} and second best in \textcolor{blue}{blue}.}
\vspace{-8pt}
\label{tab:comparison}
\resizebox{\columnwidth}{!}{
\begin{tabular}{lccc}
\toprule
& \textbf{Real-ESRGAN} & \textbf{SwinIR} & \textbf{\method} \\
\midrule
Params$\downarrow$ & 16.70M & \textcolor{blue}{11.46M} & \textcolor{red}{1.67M} \\
MACs (256$\times$256)$\downarrow$ & 1.17T & \textcolor{blue}{752.13G} & \textcolor{red}{108.28G} \\
FLOPs (256$\times$256)$\downarrow$ & 2.35T & \textcolor{blue}{1.50T} & \textcolor{red}{216.55G} \\
Inference Runtime$\downarrow$ & \textcolor{blue}{148.70\,ms} & 410.70\,ms & \textcolor{red}{83.47\,ms} \\
FPS $\uparrow$& \textcolor{blue}{6.72} & 2.44 & \textcolor{red}{11.98} \\
\midrule
PSNR $\uparrow$ & \textcolor{red}{38.1400} & 34.7282 & \textcolor{blue}{37.9102} \\
SSIM $\uparrow$ & \textcolor{red}{0.9007} & 0.8242 & \textcolor{blue}{0.8975} \\
LPIPS $\downarrow$ & \textcolor{red}{0.0974} & 0.1848 & \textcolor{blue}{0.0987} \\
DISTS $\downarrow$ & \textcolor{red}{0.0790} & 0.1344 & \textcolor{blue}{0.0837} \\
MUSIQ $\uparrow$ & \textcolor{blue}{44.2717} & 43.4717 & \textcolor{red}{44.3044} \\
BRISQUE $\downarrow$ & \textcolor{red}{34.6102} & 38.5192 & \textcolor{blue}{35.6094} \\
\bottomrule
\end{tabular}}
\vspace{-18pt}
\end{table}

\begin{table}[t]
\centering
\caption{Ablation study. Image-quality metrics
are reported on testing dataset.Best results are in \textcolor{red}{red} and second best in \textcolor{blue}{blue}.}
\vspace{-8pt}
\label{tab:ablation}
\resizebox{\columnwidth}{!}{
\begin{tabular}{lccc}
\toprule
& \textbf{Baseline} & \textbf{Lite-RRDB} & \textbf{Lite-RRDB\,+\,$\,$KD} \\
\midrule
PSNR $\uparrow$
& \textcolor{red}{38.1400} & 37.2433 & \textcolor{blue}{37.9102} \\
SSIM $\uparrow$
& \textcolor{red}{0.9007} & 0.8939 & \textcolor{blue}{0.8975} \\
LPIPS $\downarrow$
& \textcolor{red}{0.0974} & 0.1257 & \textcolor{blue}{0.0987} \\
DISTS $\downarrow$
& \textcolor{red}{0.0790} & 0.0938 & \textcolor{blue}{0.0837} \\
MUSIQ $\uparrow$
& \textcolor{blue}{44.2717} & 43.1201 & \textcolor{red}{44.3044} \\
CLIP-IQA $\uparrow$
& \textcolor{red}{0.3984} & 0.3590 & \textcolor{blue}{0.3607} \\
BRISQUE $\downarrow$
& \textcolor{red}{34.6102} & 36.9312 & \textcolor{blue}{35.6094} \\
\midrule
Params & 16.70M & \textcolor{red}{1.67M} & \textcolor{red}{1.67M} \\
MACs (256$\times$256) & 1.17T & \textcolor{red}{108.28G} & \textcolor{red}{108.28G} \\
FLOPs (256$\times$256) & 2.35T & \textcolor{red}{216.55G} & \textcolor{red}{216.55G} \\
Inference Runtime & 148.70\,ms & \textcolor{red}{83.41\,ms} & \textcolor{red}{83.47\,ms} \\
FPS & 6.72 & \textcolor{red}{11.99} & \textcolor{red}{11.98} \\
Peak Memory & 2722.67\,MB & \textcolor{red}{2667.78\,MB} & \textcolor{red}{2667.78\,MB} \\
\bottomrule
\end{tabular}}
\vspace{-12pt}
\end{table}
\subsection{Dataset and Evaluation Metrics}

\textbf{Synthetic training data.}
We use clean 400 images from DocVQA~\cite{mathew2021docvqadatasetvqadocument} and 2000 from LSDIR~\cite{LSDIR} as ground truth. Each clean
image is processed by NSP to produce noisy ones.

\textbf{Real-world testing data.}
We evaluate using real-world dataset from the Mobile
AI Denoising Dataset and MIDD~\cite{dataset2}.

\textbf{Image-quality metrics.}
We report PSNR and SSIM~\cite{wang2004image}, for which higher values indicate better
reference-based reconstruction. LPIPS~\cite{zhang2018perceptual}, and DISTS~\cite{DISTS} measure perceptual
distance, and lower values are preferred. We additionally report the
no-reference metrics MUSIQ~\cite{MUSIQ} and CLIP-IQA~\cite{CLIP-IQA}, for which higher values are
preferred, and BRISQUE~\cite{BRISQUE}, for which lower values are preferred.

\textbf{Efficiency metrics.}
We evaluate parameters, MACs, FLOPs, inference runtime, frames per
second, and peak memory. Lower parameter count, MACs,
FLOPs, runtime, and memory are preferred, while higher FPS is preferred.
\vspace{-3pt}
\subsection{Implementation Details}
\vspace{-3pt}
The Heavy Teacher is trained for
100000 iterations using
Adam optimizer with an initial learning rate of $2 \times 10^{-4}$ with a batch size of 4 and patch size of  $128\times128$.

After Teacher training, its parameters are frozen. The Student is
trained with Lite-RRDB in stage 1 and trained with knowledge distillation in stage 2 for
100000 iterations using learning rate was reduced to $1 \times 10^{-4}$.

Runtime and memory are measured with an input size of $256\times256$. 
The training process was conducted on the RTX 3090 with memory of 24 GB on the training dataset constructed by myself, consisting of both real-world noisy images and simulated noisy images.
\vspace{-3pt}
\subsection{Baselines}
\vspace{-3pt}
We compare the following models:
\vspace{-4pt}
\begin{itemize}
    \item \textbf{Heavy Teacher:} the adapted Real-ESRGAN denoising
    network.

        \item \textbf{SwinIR:} a transformer-based image restoration model.

    \item \textbf{LiteKD-Net:} our model.

\end{itemize}
\vspace{-3pt}
\subsection{Main Results}
\vspace{-3pt}
Table~\ref{tab:comparison} compares ours with SwinIR~\cite{liang2021swinir} and Real-ESRGAN~\cite{wang2021realesrgan}. It can be observed that ours reaches the greatest efficiency followed by the Real-ESRGAN~\cite{wang2021realesrgan}. Considering the image metrics, Real-ESRGAN~\cite{wang2021realesrgan} achieves the highest value in 5 metrics except MUSIQ. Though ours only achieves the second highest value in 5 metrics and the highest in MUSIQ, it's worth noticing that ours model has very close image metric value to Real-ESRGAN~\cite{wang2021realesrgan} but with great reduction in efficiency-related metrics, such as about ten-times reduction in model size, four-times reduction in MACs and FLOPs, and about two-times faster in inference speed.

\vspace{-3pt}
\subsection{Ablation Study}
\vspace{-3pt}
We perform an ablation study to isolate the effects of Lite-RRDB and Noise-KD. It can be seen that Lite-RRDB has a great reduction in model size and faster inference speed, but with a bit reduction in imgae quality. However, Noise-KD helps improve image quality a lot, very close to the baseline.

\vspace{-3pt}
\subsection{Qualitative Analysis}
\begin{figure}[t]
    \centering
    \includegraphics[width=\columnwidth]
    {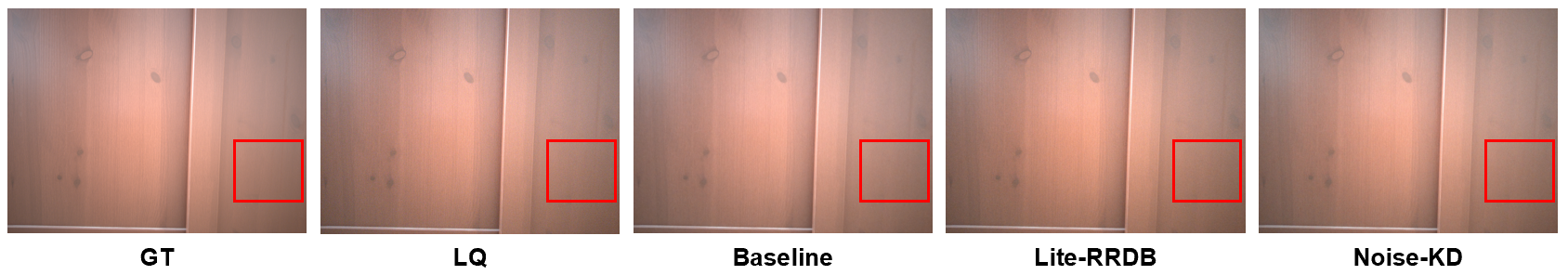}
    \vspace{-12pt}
    \caption{Qualitative comparison. From left to right:
    GT, LQ, Baseline, Lite-RRDB,
    and Lite-RRDB + Noise-KD.}
    \label{fig:qualitative}
    \vspace{-18pt}
\end{figure}

\vspace{-3pt}
Figure~\ref{fig:qualitative} presents representative visual comparisons
on real-world noisy images. The Heavy Teacher effectively suppresses noise
and preserves fine structures, but requires substantially more
computation. Lite-RRDB produces faster results but may remove subtle
textures or leave residual noise in difficult regions. After knowledge
distillation, the proposed model recovers sharper boundaries and more
natural textures while retaining the efficiency of the lightweight
architecture. More comparison images can be seen in Appendix B.

\vspace{-3pt}
\subsection{Limitations}
\vspace{-3pt}
The simulation pipeline operates on approximately linearized sRGB
images rather than true camera RAW. Therefore, it does not
model the complete camera imaging pipeline, including demosaicing,
white balance, color correction, tone mapping, or camera-specific
nonlinear processing.

\section{Conclusion}
\label{sec:conclusion}

This paper presented LiteKD-Net, a physics-guided lightweight
knowledge-distilled network for mobile image denoising. NSP generates noise in linear RGB domain. A high-capacity RRDB Teacher is adapted from Real-ESRGAN by
removing all upsampling modules, while a lightweight Student replaces
standard convolutions with Lite-RRDB blocks.
Feature-level knowledge distillation transfers intermediate restoration
knowledge without increasing computation cost.

Experimental results show that ours reaches good trade-off in computational efficiency and image quality.
Compared with our baseline, we reduces the model size by ten times and increase the inference speed by two times with close restoration quality. 

\bibliographystyle{IEEEtran}
\bibliography{references}

\clearpage
\appendices

\section{Detailed Implementation of the Noise Simulation Pipeline}
\label{app:nsp_details}

This appendix provides the detailed implementation of the proposed
Noise Simulation Pipeline (NSP). The pipeline is adapted from the
physics-based RAW noise formation model of Wei et
al.~\cite{wei2020physics}, which models photon shot noise, read noise,
row noise, and quantization noise. In contrast to the original
camera-calibrated RAW formulation, our implementation operates on
approximately linearized sRGB images and additionally introduces pixel
crosstalk to model local spatial correlation between neighboring pixels.

Let
$\mathbf{y}_{s}\in[0,1]^{3\times H\times W}$
denote a clean sRGB image. The corresponding simulated noisy image is
denoted by
$\hat{\mathbf{x}}_{s}$.
All noise components are applied in an approximately linear RGB domain.

\subsection{Approximate sRGB Linearization}
\label{app:gamma}

The input image is first normalized from $[0,255]$ to $[0,1]$ and
approximately converted from sRGB to linear RGB using gamma expansion:

\begin{equation}
\label{eq:app_gamma_expand}
\mathbf{y}_{0}
=
\Gamma(\mathbf{y}_{s})
=
\mathbf{y}_{s}^{\gamma},
\qquad
\gamma=2.2.
\end{equation}

This approximation removes the nonlinear display encoding so that the
following sensor-noise operations are applied to values that more
closely represent physical signal intensity.

\subsection{Photon Shot Noise}
\label{app:shot_noise}

Photon shot noise represents the random arrival of photons during
exposure. Given the effective system-gain parameter $K$, we first
compute the expected photon count:

\begin{equation}
\label{eq:app_photon_count}
\boldsymbol{\mu}_{p}
=
\frac{\mathbf{y}_{0}}{K}.
\end{equation}

A Poisson-distributed photon count is then sampled independently for
each pixel and color channel:

\begin{equation}
\label{eq:app_poisson_sample}
\mathbf{p}
\sim
\operatorname{Poisson}
\left(
\boldsymbol{\mu}_{p}
\right).
\end{equation}

The signal is converted back to the normalized image domain as

\begin{equation}
\label{eq:app_shot_signal}
\mathbf{y}_{1}
=
K\mathbf{p}
=
K\operatorname{Poisson}
\left(
\frac{\mathbf{y}_{0}}{K}
\right).
\end{equation}

Its conditional expectation and variance are

\begin{equation}
\label{eq:app_shot_statistics}
\mathbb{E}
\left[
\mathbf{y}_{1}
\mid
\mathbf{y}_{0}
\right]
=
\mathbf{y}_{0},
\qquad
\operatorname{Var}
\left[
\mathbf{y}_{1}
\mid
\mathbf{y}_{0}
\right]
=
K\mathbf{y}_{0}.
\end{equation}

Therefore, photon shot noise is signal-dependent.

\subsection{Tukey--Lambda Read Noise}
\label{app:read_noise}

Electronic read noise may exhibit a heavy-tailed distribution. Following
Wei et al.~\cite{wei2020physics}, we model it using the Tukey--Lambda
distribution.

For each pixel and color channel, we independently sample

\begin{equation}
\label{eq:app_uniform_read}
u(c,h,w)
\sim
\mathcal{U}(0,1).
\end{equation}

For numerical stability, the sampled values are clipped to

\begin{equation}
\label{eq:app_uniform_clamp}
u
\leftarrow
\operatorname{clip}
\left(
u,\epsilon,1-\epsilon
\right),
\qquad
\epsilon=10^{-6}.
\end{equation}

The Tukey--Lambda read noise is generated using its inverse cumulative
form:

\begin{equation}
\label{eq:app_tukey_lambda}
\mathbf{n}_{\mathrm{read}}
=
\sigma_{\mathrm{TL}}
\frac{
u^{\lambda}
-
(1-u)^{\lambda}
}{
\lambda
},
\end{equation}

where $\lambda$ controls the distribution shape and
$\sigma_{\mathrm{TL}}$ controls its scale. The resulting signal is

\begin{equation}
\label{eq:app_after_read}
\mathbf{y}_{2}
=
\mathbf{y}_{1}
+
\mathbf{n}_{\mathrm{read}}.
\end{equation}

\subsection{Row Noise}
\label{app:row_noise}

Row noise models horizontal banding caused by row-wise sensor readout.
For each row $h$, one Gaussian offset is sampled:

\begin{equation}
\label{eq:app_row_sample}
r_h
\sim
\mathcal{N}
\left(
0,\sigma_r^2
\right).
\end{equation}

The same offset is broadcast across all columns and all three color
channels:

\begin{equation}
\label{eq:app_row_broadcast}
\mathbf{n}_{\mathrm{row}}(c,h,w)
=
r_h,
\qquad
c\in\{1,2,3\}.
\end{equation}

The signal after row-noise simulation becomes

\begin{equation}
\label{eq:app_after_row}
\mathbf{y}_{3}
=
\mathbf{y}_{2}
+
\mathbf{n}_{\mathrm{row}}.
\end{equation}

\subsection{Pixel Crosstalk}
\label{app:pixel_crosstalk}

Our main extension to the original noise model is pixel crosstalk.
Instead of assuming that neighboring sensor pixels are spatially
independent, we allow a small fraction of the local signal to diffuse
to adjacent pixels.

The proposed crosstalk kernel is

\begin{equation}
\label{eq:app_crosstalk_kernel}
\boldsymbol{\kappa}_{\mathrm{xt}}
=
\begin{bmatrix}
\rho/4 & \rho & \rho/4 \\
\rho & 1-5\rho & \rho \\
\rho/4 & \rho & \rho/4
\end{bmatrix},
\end{equation}

where $\rho$ is the crosstalk strength. The kernel satisfies

\begin{equation}
\label{eq:app_kernel_sum}
\sum_{i=-1}^{1}
\sum_{j=-1}^{1}
\boldsymbol{\kappa}_{\mathrm{xt}}(i,j)
=
1,
\end{equation}

so it approximately preserves the local mean intensity.

The kernel is independently applied to each color channel:

\begin{equation}
\label{eq:app_crosstalk_operation}
\mathbf{y}_{4}^{(c)}
=
\mathbf{y}_{3}^{(c)}
*
\boldsymbol{\kappa}_{\mathrm{xt}},
\qquad
c\in\{1,2,3\},
\end{equation}

where $*$ denotes two-dimensional spatial convolution. In the
implementation, this operation is performed using grouped convolution
with the number of groups equal to the number of channels. Thus, the operation introduces spatial interaction between neighboring
pixels without directly mixing the RGB channels. A padding size of one
pixel is used to preserve the original spatial resolution.

With the default setting $\rho=0.03$, the kernel becomes

\begin{equation}
\label{eq:app_crosstalk_numeric}
\boldsymbol{\kappa}_{\mathrm{xt}}
=
\begin{bmatrix}
0.0075 & 0.03 & 0.0075 \\
0.03 & 0.85 & 0.03 \\
0.0075 & 0.03 & 0.0075
\end{bmatrix}.
\end{equation}

The central pixel remains dominant, while $15\%$ of the total kernel
weight is distributed to neighboring pixels. This produces mild local
spatial correlation.

\subsection{Quantization Noise}
\label{app:quantization_noise}

The analog-to-digital conversion stage is approximated using uniform
quantization noise. For a $B$-bit signal, the quantization step is

\begin{equation}
\label{eq:app_quant_step}
q
=
\frac{1}{2^{B}-1}.
\end{equation}

Let

\begin{equation}
\label{eq:app_quant_uniform}
u_q(c,h,w)
\sim
\mathcal{U}(0,1).
\end{equation}

The quantization noise is generated as

\begin{equation}
\label{eq:app_quant_noise}
\mathbf{n}_{q}
=
\left(
u_q-\frac{1}{2}
\right)q,
\end{equation}

which is equivalent to

\begin{equation}
\label{eq:app_quant_distribution}
\mathbf{n}_{q}
\sim
\mathcal{U}
\left(
-\frac{q}{2},
\frac{q}{2}
\right).
\end{equation}

Our implementation uses $B=14$, giving

\begin{equation}
\label{eq:app_quant_14bit}
q
=
\frac{1}{2^{14}-1}
=
\frac{1}{16383}.
\end{equation}

The signal after quantization-noise simulation is

\begin{equation}
\label{eq:app_after_quant}
\mathbf{y}_{5}
=
\mathbf{y}_{4}
+
\mathbf{n}_{q}.
\end{equation}

\subsection{Clipping and Conversion Back to sRGB}
\label{app:output_conversion}

The simulated linear-domain image is clipped to the valid intensity
range:

\begin{equation}
\label{eq:app_linear_output}
\hat{\mathbf{x}}_{\mathrm{lin}}
=
\operatorname{clip}
\left(
\mathbf{y}_{5},
0,1
\right).
\end{equation}

Finally, inverse gamma transformation converts the result back to sRGB:

\begin{equation}
\label{eq:app_inverse_gamma}
\hat{\mathbf{x}}_{s}
=
\Gamma^{-1}
\left(
\hat{\mathbf{x}}_{\mathrm{lin}}
\right)
=
\hat{\mathbf{x}}_{\mathrm{lin}}^{1/\gamma},
\qquad
\gamma=2.2.
\end{equation}

The output is converted to an unsigned 8-bit image and stored using the
same filename as its clean ground-truth counterpart, thereby preserving
strict pixel-level alignment between the LQ and GT images.

\subsection{Complete NSP Formulation}
\label{app:complete_nsp}

Combining all operations, the complete NSP can be written as

\begin{equation}
\label{eq:app_complete_nsp}
\boxed{
\begin{aligned}
\hat{\mathbf{x}}_{s}
&= \Gamma^{-1} \Bigg[ \\
&\quad \operatorname{clip}
\left(
\mathcal{C}_{\rho}
\left[
K\operatorname{Poisson}
\left(
\frac{\Gamma(\mathbf{y}_{s})}{K}
\right)
+ \mathbf{n}_{\mathrm{read}}
+ \mathbf{n}_{\mathrm{row}}
\right]
+ \mathbf{n}_{q},
0,
1
\right)
\Bigg]
\end{aligned}
}
\end{equation}

where $\mathcal{C}_{\rho}(\cdot)$ denotes channel-wise convolution with
the proposed pixel-crosstalk kernel in
Eq.~\eqref{eq:app_crosstalk_kernel}.

\subsection{Parameter Settings}
\label{app:nsp_parameters}

The fixed parameters used in our implementation are summarized in
Table~\ref{tab:nsp_parameters}.

\begin{table}[!t]
\centering
\caption{Parameter settings of the proposed NSP.}
\label{tab:nsp_parameters}
\begin{tabular}{ccl}
\hline
Symbol & Value & Description \\
\hline
$\gamma$ & $2.2$ & Gamma exponent \\
$K$ & $0.01$ & Effective shot-noise parameter \\
$\lambda$ & $-0.1$ & Tukey--Lambda shape \\
$\sigma_{\mathrm{TL}}$ & $0.02$ & Read-noise scale \\
$\sigma_r$ & $0.01$ & Row-noise standard deviation \\
$\rho$ & $0.03$ & Pixel-crosstalk strength \\
$B$ & $14$ & Quantization bit depth \\
$\epsilon$ & $10^{-6}$ & Numerical clipping constant \\
\hline
\end{tabular}
\end{table}

Unlike the camera-specific calibration procedure in
Wei et al.~\cite{wei2020physics}, these parameters are fixed for all
training images. Therefore, our NSP should be regarded as a
physics-guided RGB noise approximation with explicit pixel-crosstalk
modeling, rather than a camera-calibrated RAW sensor simulator.

\section{Additional Results}
\label{app:additional_results}

\begin{figure}[htbp]
    \centering
    \includegraphics[width=\columnwidth]
    {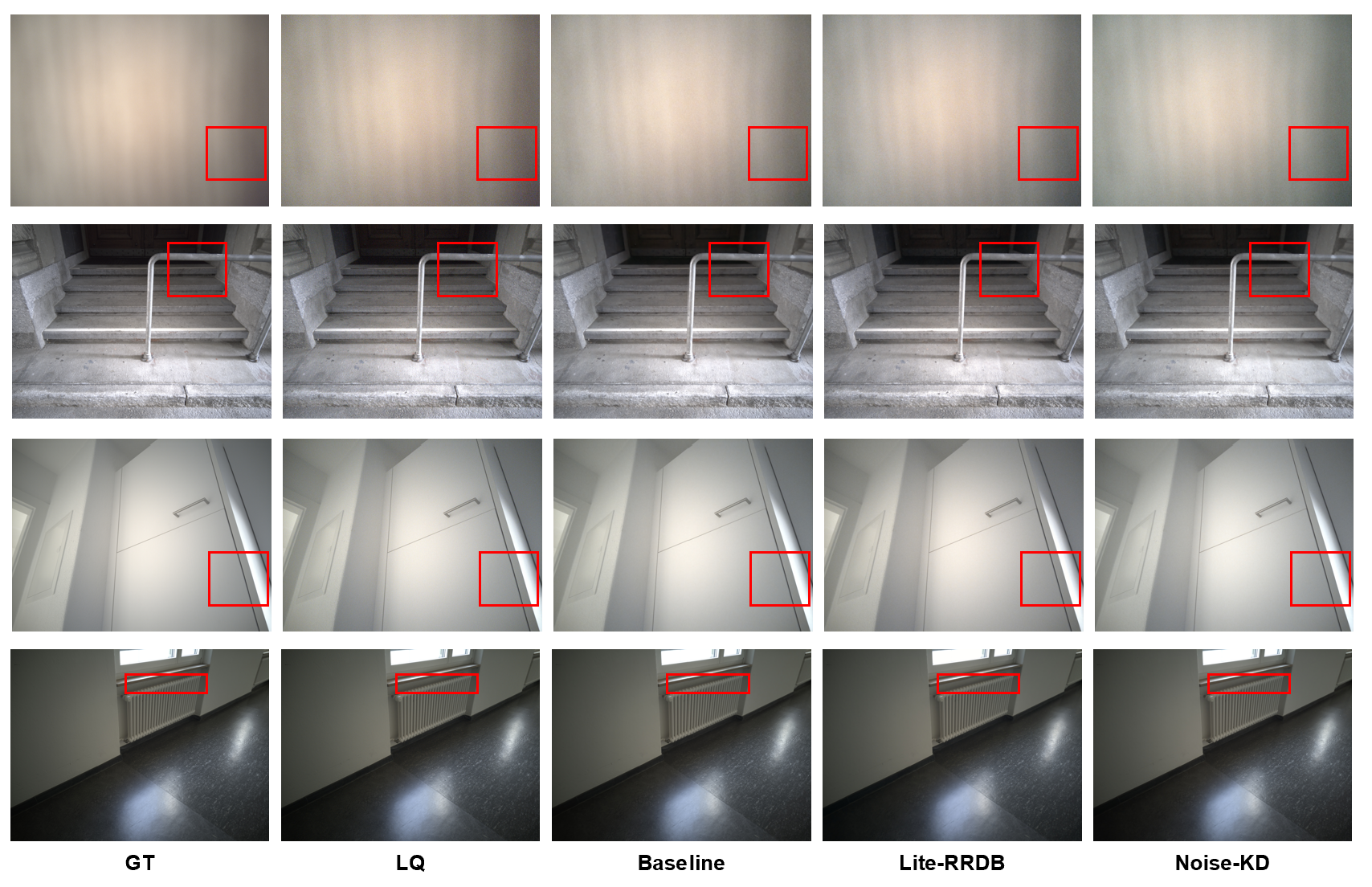}
    \vspace{-20pt}
    \caption{Qualitative comparison. From left to right:
    GT, LQ, Baseline, Lite-RRDB,
    and Lite-RRDB + Noise-KD.}
    \label{more_comp}
\end{figure}

Extra comparison are shown in the Fig.~\ref{more_comp}.
\vspace{-10pt}

\end{document}